\documentclass[runningheads]{llncs}

\usepackage[utf8]{inputenc}
\usepackage{graphicx}
\usepackage{todonotes}
\usepackage[ruled]{algorithm2e}
\usepackage{subcaption}
\usepackage{newtxtext,newtxmath}
\usepackage{mathtools}
\usepackage{hyperref}
\usepackage{bm}

\allowdisplaybreaks

\title{Non-Parametric Calibration of Probabilistic Regression}

\author{Hao Song\inst{1} \and Meelis Kull\inst{2} \and Peter Flach\inst{1}}

\institute{Intelligent Systems Laboratory, University of Bristol, Bristol, UK\\ 
\email{\{Hao.Song,Peter.Flach\}@bristol.ac.uk} 
\and University of Tartu, Tartu, Estonia\\\email{meelis.kull@ut.ee}
}

\begin{document}

\maketitle


\begin{abstract}
The task of calibration is to retrospectively adjust the outputs from a machine learning model to provide better probability estimates on the target variable.
While calibration has been investigated thoroughly in classification, it has not yet been well-established for regression tasks.
This paper considers the problem of calibrating a probabilistic regression model to improve the estimated probability densities over the real-valued targets.
We propose to calibrate a regression model through the cumulative probability density, which can be derived from calibrating a multi-class classifier. 
We provide three non-parametric approaches to solve the problem, two of which provide empirical estimates and the third providing smooth density estimates.
The proposed approaches are experimentally evaluated to show their ability to improve the performance of regression models on the predictive likelihood.
\end{abstract}

\section{Introduction}
\label{sec:intro}

In predictive machine learning, (probability) calibration refers to a set of techniques that applies post-hoc modelling to correct the outputs from trained classifiers, so that the final outputs are better probability estimates on the target variable. 
Given a probabilistic two-class classifier, an output $s \in [0,1]$ for the positive class is calibrated if the following condition holds: for all the instances receiving this prediction value of $s$, the probability of observing a positive label is $s$.
From a frequentist point of view, a $0.5$ estimated probability of rain tomorrow is calibrated if, among all the days receiving this probability estimate, half of those days it indeed rained.
A Bayesian might say that this $0.5$ is calibrated if it covers a group of days about which we should be maximally uncertain whether it will rain or not.

Calibration helps to make optimal decisions (e.g. setting a threshold on the classifier's score) in cost-sensitive classification \cite{Zadrozny2002} and allows adapting to changing cost parameters without re-training the model. 
However, being calibrated does not necessarily imply that the classifier has good performance.
For instance, a constant classifier outputting the marginal target distribution is calibrated by definition, but it is not a good predictive classifier as it does not separate the classes.
In general, given a trained uncalibrated classifier, applying calibration can help improve the estimated probabilities, but not to further separate the feature space.  
%
One of the well-known parametric approaches is Logistic calibration 
\cite{Platt1999}, which uses the logistic function to map the SVM margins into better calibrated probabilities.
\cite{Kull2017} proposed the Beta calibration approach that allows more flexible adjustment besides a simple Sigmoid shape. 

\begin{figure}[t]
    \centering
    \includegraphics[width=0.32\textwidth]{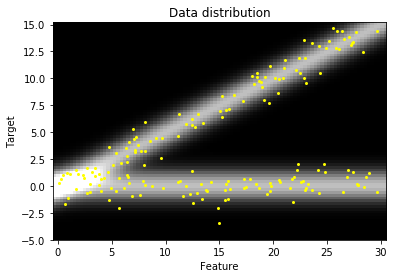}
    \includegraphics[width=0.32\textwidth]{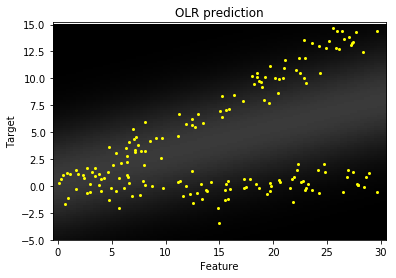}
    \includegraphics[width=0.32\textwidth]{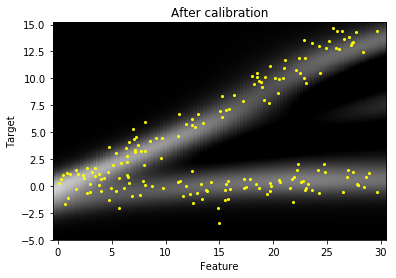}
    \caption{The left figure shows the true conditional distribution of the target given the feature (in greyscale), and some data points drawn from the distribution (in yellow). The middle figure shows the densities predicted from the Ordinary Least Squares regression method (unimodal Gaussians). The results of calibrating these predicted densities using our proposed GPC calibration method have been shown in the right figure.}
    \label{fig:toy_example}
\end{figure}

Calibration can be beneficial for regression tasks as well.
Consider the toy dataset with a univariate regression task in Figure~\ref{fig:toy_example}. 
In this task the feature contains useful information only in about half of the data (instances on the ascending line) while being non-informative on the other half (instances on the flat line). 
The actual conditional density of the target variable given the feature is shown in the left figure using background colour. 
The middle figure shows the predicted conditional densities from Ordinary Least Squares regression algorithm, which can be interpreted as assuming Gaussian conditional distributions with shared standard deviation. 
Clearly, the OLS doesn't capture the shape of the distribution, and considerably over/under-estimates the densities around some regions.
While this can be fixed by applying a different model that suits the data distribution, it requires to know or assume the true distribution family, which is often not feasible for many high dimensional and complex datasets.
Alternatively, one can adopt methods from the field of conditional density estimation \cite{Schapire2002,Bishop2006,Sugiyama2010} or quantile regression \cite{Koenker2001,Li2007}, without knowing the exact parametric form of the target distribution.
However, such approaches normally require additional kennels or basis functions to be selected, which can be problematic for certain feature spaces.

In this paper we instead propose to take the predicted densities and improve them using a calibration procedure.
The proposed approaches enable us to directly take the outputs from the well-founded regression models, and apply post-hoc calibrations to obtained better conditional density estimations. 
The figure on the right shows the result from applying GPC calibration on the OLS model (one of the calibration methods proposed in this paper).

There are several benefits from calibrating a regression model: 
(1) As in the case of classification, calibrating a regression model improves the probability estimates on the target variable, and hence reduces the uncertainties in decision making.
(2) Calibration can help to correct a mismatched distribution assumption, e.g.\ when the regression model assumes that the residuals are Gaussian whereas actually they are not.
%
On the other hand, calibration of regression models can be hard:
(1) Existing calibration methods from classification cannot be directly applied, as the conditional distribution of the target variable given the features is continuous rather than categorical. 
(2) Simple parametric methods might be insufficient to model the richness of continuous distributions.

In this paper we generalise the concept of calibration to regression tasks.
We contribute to the problem by first defining what calibrated regression is, and then demonstrate the relationship between calibrating density functions and cumulative distribution functions.
Based on the theory, an empirical approach is proposed to adapt the existing framework of calibration for classification.
Another approach is further proposed to provide smooth estimation on the densities, which is based on a Gaussian process classifier.
The calibrated density functions are non-parametric and hence suitable for any potential target distributions.
We experimentally show that our calibration method can indeed increase the performance of regression models on their estimated densities.

The structure of the paper is as follows. 
Section~\ref{sec:binary_cal} introduces calibration of binary classifiers.
Section~\ref{sec:reg_cal} gives the definitions of calibration and empirical calibration of a regression, with further theorems on the link between calibrated density functions and calibrated cumulative distribution functions.
Section~\ref{sec:e_cal} introduces a simple method to adopt existing binary approaches to calibrate a regression empirically.
Section~\ref{sec:gp_cal} shows the proposed non-parametric approach based on a Gaussian Process classifier.
Section~\ref{sec:exp} shows the experiments and results and Section~\ref{sec:con} concludes the paper.

\section{Classifier Calibration}
\label{sec:binary_cal}

In this section we introduce the general concept of calibration in binary classification. 
%
We use $\mathbf{X}$ (in space $\mathbb{X}$) and $Y$ (in space $\mathbb{Y}$) to represents the random variable for the feature vector and target value, respectively.
In the case of $K$-class ($K \geq 2$) classification, we denote $\mathbb{Y} = \{1,...,K\}$.
The small case $\mathbf{x}$ and $y$ are used to denote an instance of the feature vector and the target.  
Additionally, we use the notation $\mathbb{P}$ and $p$ to distinguish probability mass and probability density.

A probabilistic classifier is defined as a function $f: \mathbb{X} \rightarrow (\mathbb{Y} \rightarrow [0,1])$, so that:
\small
\begin{align}
    f_{(\mathbf{x})}(Y) = \hat{\mathbb{P}}(Y \mid \mathbf{X}=\mathbf{x})
\end{align}
\normalsize
Hence, the model can take a feature vector $\mathbf{x}$ and outputs a estimated probability mass function (e.g. categorical likelihood) on the target variable $Y$. 

\subsection{Binary Classification}
With the notations above, we can now give the definition of calibration in binary classification.

\begin{definition}[Calibrated Binary Classification]
A classifier $f$ is defined as calibrated if and only if, given $\forall s \in [0,1]$, the following holds:
\begin{align}
    \mathbb{P}\Big(Y=1 \mid  f_{(\mathbf{X})}(Y=1) = s\Big) = s
\end{align}
\end{definition}
%

As discussed in previous work \cite{Kull2017}, even if a classifier shows good performance on metrics such as accuracy or F-score, its output might not be calibrated. 
Therefore, depending on the properties of the classifier, we have two kinds of calibration.
The first kind arises when the model doesn't provide probability estimates on the target variable, and we therefore need to derive calibrated probabilities from its outputs.
For instance, a SVM by default only predicts the margins optimised with hinge loss, and therefore requires calibration for probabilistic outputs.
The second kind occurs when a model is already probabilistic, as the results might not be accurate due to its assumptions or approximations, which also requires to be further calibrated to generate better results.
One example in this category is the Naive Bayes classifier. 
While it is probabilistic and optimised via maximum likelihood, the independence assumption among the features makes its outputs poorly calibrated in general.

To solve the issues above, different approaches have been introduced to post-calibrate such classifiers.
In this paper we only focus on the second kind of calibration, and we define binary calibration as a function $c: [0,1] \rightarrow [0,1]$, so that, for each feature vector $\mathbf{x}$, a calibration is able to provide an estimation $c\Big(f_{(\mathbf{X}=\mathbf{x})}(Y=1)\Big)$ on the true calibrated score $\mathbb{P} \Big( Y=1 \mid f_{(\mathbf{X}=\mathbf{x})}(Y=1)\Big)$.
In \cite{Zadrozny2002,Cohen2004}, the authors provide a list of properties and benefits by having a calibrated output.
Empirical binning has been used as a baseline method, which estimates an empirical distribution on the predicted score \cite{Zadrozny2001}.   
Recently, \cite{Naeini2015} proposed a Bayesian binning approach to improve the estimates by performing inference on a hidden binning scheme.
Isotonic regression with its related PAV algorithm is one of the major non-parametric calibration methods \cite{Zadrozny2002,Fawcett2007}.
The method calibrates a model by recursively averaging neighbouring non-monotonic scores, so that a piece-wise constant non-decreasing calibration map is obtained at the end.  


\textit{Logistic calibration} can be seen as a special case of 1-D Logistic regression, where the input is the uncalibrated output, 
and the model is fitted with the target $Y$ to predict the calibrated probabilities. 
In the case of a probabilistic model, denoting $s = f_{(\mathbf{X}=\mathbf{x})}(Y=1)$, logistic calibration is given as:
\small
\begin{align}
    c_{\text{Logistic}} (s) = \frac{1}{1 + \left(\exp(\gamma \cdot s + \delta)\right)^{-1}}
\end{align}
\normalsize
Here $\gamma$, $\delta \in \mathbb{R}$ are the estimated parameters.

One way to interpret logistic calibration, or in general multivariate logistic regression, is through the Linear Discriminant Analysis (LDA) and the corresponding Gaussian assumption \cite{Flach2012,Murphy2012}.
%
%
%
%
%
%
Mathematically, LDA and logistic regression share the same function while calculating the target distribution with a given feature.
The difference is that, while LDA estimates the parameters as the class prior, Gaussian means, and shared covariance matrix, logistic regression directly fits the parameters $\gamma$ and $\delta$ through numerical optimisation.


While the Gaussian assumption is reasonable for a input defined in $\mathbb{R}$, it becomes less appropriate while calibrating a probabilistic model, where the input is instead in the interval $[0, 1]$.
Kull et al then propose to instead use the Beta distribution to model the conditional probability $p(S|Y)$, with $S = f_{\mathbf{X}}(Y=1)$ denoting the random variable for the score produced by a classifier:
\small
\begin{align*}
    p(S=s|Y=y) = \frac{s^{\alpha_y-1}(1-s)^{\beta_y-1}}{B(\alpha_y,\beta_y)}
\end{align*}
\normalsize
With $B(\alpha_y,\beta_y)$ denoting the Beta normalisation constant.
\textit{Beta calibration} can then be stated in the following form:
\small
\begin{align*}
   c_{beta}(s) = \frac{1}{1 + \left(\exp(m)\frac{s^{a}}{(1-s)^{b}}\right)^{-1}}
\end{align*}
\normalsize
Here $a = \alpha_{1} - \alpha_{2}$, $b = \beta_{2} - \beta_{1}$, and $m = log\frac{\mathbb{P}(Y=1)}{\mathbb{P}(Y=2)} + log \frac{B(\alpha_2, \beta_2)}{B(\alpha_1, \beta_1)}$.

As in the case of LDA and logistic regression, Beta calibration can also be fitted via a generative approach or a discriminative approach.
In the discriminative case, the authors also show that the parameters of a Beta calibration can be fitted through logistic regression with a bi-variate input $z = [\ln s, -\ln (1-s)]$. 
Beta calibration improves logistic regression as the calibration map is not necessarily sigmoidal, and hence more versatile for the general purpose of calibration.
The experiments in \cite{Kull2017,Kull2017b} shows improvements on log-loss and Brier score of Beta calibration over other calibration approaches on a set of model classes, including Naive Bayes, Logistic Regression, Support Vector Machine (SVM), Random Forest (RF) and Multi-Layer Perceptron (MLP), and two variants of AdaBoost. 

\subsection{Multi-class Classification}

The concept of calibration can be generalised to multi-class classification as follows.

\begin{definition}[Calibrated Multi-class Classification]
\label{def:mul_cal}
Let $f$ be a classifier, denoting $\mathbf{S} = [f_{(\mathbf{X})}(Y=1), \dots, f_{(\mathbf{X})}(Y=K)]$ as the random variable for the predicted probability vector, we define $f$ to be calibrated if and only if, for every possible vector $\mathbf{s}=[s_1, \dots ,s_K]$ in the $K$-dimensional probability simplex $\{s_i \in [0,1]$, $\sum_{i=1}^{N}s_i = 1\}$, the following holds:
\begin{align}
\label{eq:mul_cal}
    \mathbb{P}\Big(Y=i \mid \mathbf{S} = \mathbf{s} \Big) &= s_i 
\end{align}
\end{definition}

Therefore, multi-class calibration asks that, given a predicted $K$-dimensional probability vector, every dimension of the vector is calibrated with the corresponding target class.
The simplest approach to calibrate a multi-class classifier is to apply binary calibration on each target value with the one-vs-rest strategy, and eventually normalised obtained probability vector. 
As an direct extension, multinomial logistic regression is also commonly used in this case.

Empirical binning has been used as a baseline method, which estimates an empirical distribution on the predicted score \cite{Zadrozny2001}.   
Recently, \cite{Naeini2015} proposed a Bayesian binning approach to improve the estimates by performing inference on a hidden binning scheme.
Isotonic regression with its related PAV algorithm is one of the major non-parametric calibration methods \cite{Zadrozny2002,Fawcett2007}.
The method calibrates a model by recursively averaging neighbouring non-monotonic scores, so that a piece-wise constant non-decreasing calibration map is obtained at the end.  


\section{Calibration of Probabilistic Regression}
\label{sec:reg_cal}

Now we move on to probabilistic regression models, where $\mathbb{Y} = \mathbb{R}$.
We again define a probabilistic regression as a function, denoted as $g: \mathbb{X} \rightarrow \mathbb{S}$, 
with 
$$\mathbb{S} = \{ \mathsf{s}: \mathbb{R} \rightarrow [0, +\infty) \mid \int_{-\infty}^{+\infty} \mathsf{s}(y) \: dy = 1\}$$
In words, a trained probabilistic regression model provides a probability density function $g_{(\mathbf{x})}(Y)$ on the target variable $Y$ given a feature vector $\mathbf{x}$.

As shown above, calibration is a property of the predicted probability mass. 
To define calibration in probabilistic regression, we therefore need to consider the integral of density functions.
One common approach to generalise concepts from discrete to continuous, as in deriving limiting density of discrete points from Shannon Entropy or certain conditional density estimation approaches \cite{Frank2009}, is to apply binning on the continuous variable. 
To begin with, consider a multi-class scenario, given a set of $K+2$ values, $(t_0 = -\infty) < t_1 < \dots < t_K < (t_{K+1} = +\infty)$, 
according to Definition~\ref{def:mul_cal}, a probabilistic regression model is calibrated on these values if the following equation is satisfied:
\begin{align}
\label{eq:reg_multi}
    \mathbb{P}\Big(t_{i-1} < Y \leq t_{i} \mid \int_{t_0}^{t_1}g_{(\mathbf{X})}(y) \: dy = s_1, \dots,
    \int_{t_K}^{t_{K+1}}g_{(\mathbf{X})}(y) \: dy = s_{K+1}\Big) = s_i
\end{align}
We hence denote $G: \mathbb{X} \rightarrow \mathbb{Q}$, with $\mathbb{Q} = \{ \mathsf{q} \mid \mathsf{q}: \mathbb{R} \rightarrow [0,1], \mathsf{q}(t) = \int_{y=-\infty}^{t} \mathsf{s}(y) \: dy, \mathsf{s} \in \mathbb{S}\}$.
Therefore, given a pair of $(\mathbf{x}, t)$, $G$ can provide a estimated probability mass:
\begin{align}
    G_{(\mathbf{x})}(t) = \hat{\mathbb{P}}(Y \leq t | \mathbf{X} = \mathbf{x}) 
\end{align}
Equation~\ref{eq:reg_multi} can then be turned into the following definition via denoting $q_i = \sum_{j=1}^{i} s_j$:
\begin{definition}[Empirically Calibrated Probabilistic Regression]
\label{def:e_cal}
Denoting $g$ and $G$ as above, a probabilistic regression $g$ is said to be empirically calibrated on  $(t_0 = -\infty) < t_1 < \dots < t_K < (t_{K+1} = +\infty)$, if for $\forall (0 \leq q_1 < \dots < q_K \leq 1)$, the following equation holds:
\begin{align}
    \mathbb{P}(Y \leq t_i \mid G_{(\mathbf{X})}(t_1) = q_1, \dots, G_{(\mathbf{X})}(t_K) = q_K) = q_i
\end{align}
\end{definition}
Here we omit $t_0 = -\infty$ and $t_{K+1} = \infty$, as by definition we always have $G_{(\mathbf{X})}(-\infty) = 0$ and $G_{(\mathbf{X})}(\infty) = 1$.

It then makes sense to define calibrated regression as the limiting case where $K \to +\infty$ and $t_{i+1} - t_{i} \to 0$, that is, we have a set of infinitely smooth values of $t$.
Therefore, the condition $(G_{\mathbf{X}}(t_1) = q_1, \dots, G_{\mathbf{X}}(t_K) = q_K)$ can be replaced with $ (G_{\mathbf{X}} = \mathsf{q})$, with $\mathsf{q}$ being an instance of a cumulative distribution function in $\mathbb{Q}$.
This leads to our definition of a calibrated probabilistic regression:
\begin{definition}[Calibrated Probabilistic Regression]
\label{def:reg_cal}
With $g$ and $G$ as defined above, a probabilistic regression $g$ is said to be calibrated if for $\forall t \in \mathbb{R}$, $\mathsf{q} \in \mathbb{Q}$, the following equation holds:
\begin{align}
    \mathbb{P}(Y \leq t \mid G_{(\mathbf{X})} = \mathsf{q}) = \mathsf{q}(t)
\end{align}
\end{definition}

While the definition above is formalised through the predicted cumulative distribution in analogy with classification, we now show that, being calibrated in this sense also leads to ``calibrated'' densities.
\begin{lemma}
If a regression model $g$ is calibrated, as defined in Definition~\ref{def:reg_cal}, then for $\forall t \in \mathbb{R}$, $\mathsf{s} \in \mathbb{S}$ the following holds:
%
\begin{align}
    p(Y=t \mid g_{(\mathbf{X})} = \mathsf{s}) = \mathsf{s}(t)
\end{align}
\begin{proof}
Denoting $g$ and $G$ as above, $\mathsf{s} \in \mathbb{S}$, $\mathsf{q} \in \mathbb{Q}$ as a particular pair of PDF and CDF, so that $\mathsf{q}(t) = \int_{-\infty}^{t} \mathsf{s}(y) \: dy$, we have:
\begin{align*}
    \mathbb{X}_{(\mathsf{q})} &= \{\mathbf{x} \mid G_{(\mathbf{x})} = \mathsf{q} \}\\
    & = \{\mathbf{x} \mid g_{(\mathbf{x})} = \mathsf{s} \} 
\end{align*}
Hence:
\begin{align*}
    \mathbb{P}(Y \leq t \mid G_{(\mathbf{X})} = \mathsf{q}) &= \mathbb{P}(Y \leq t \mid \mathbf{X} \in \mathbb{X}_{(\mathsf{q})})\\
    &= \mathbb{P}(Y \leq t \mid g_{(\mathbf{x})} = \mathsf{s})
\end{align*}
Now we can show:
\begin{align*}
    p(Y = t \mid g_{(\mathbf{x})} = \mathsf{s}) &= \lim_{\Delta \to 0} \frac{\mathbb{P}(Y \leq t + \Delta \mid g_{(\mathbf{x})} = \mathsf{s}) - \mathbb{P}(Y \leq t \mid g_{(\mathbf{x})} = \mathsf{s})}{\Delta} \\
    &= \lim_{\Delta \to 0} \frac{\mathsf{q}(t+\Delta) - \mathsf{q}(t)}{\Delta}\\
    &= \mathsf{s}(t)
\end{align*}
\end{proof}
\end{lemma}

A important consequence of this lemma is: calibrating a regression with its predicted cumulative distributions can also improve the estimated densities on the target, which means that for all the instances receiving a prediction of $\mathsf{s}$, the PDF of $Y$ is $\mathsf{s}$.
Therefore, we can use the log-likelihood of the predicted PDFs as a measure to examine whether a model is well calibrated.

We are finally in a position to define calibration of a regression model as a function $c: \mathbb{R} \times \mathbb{Q} \rightarrow [0, 1]$, which takes a target value $t$ and a predicted CDF $\mathsf{q}$, and outputs a calibrated probability for $Y \leq t$ given $\mathsf{q}$.
However, there is one major difficulty to design such post-calibration approaches: the input space of calibration is a set of functions.
In classification, as the inputs are probability vectors, it is simple to adopt certain existing models, such as logistic regression.
The situation is even simpler for a binary case, where the input space is the interval $[0, 1]$, which supports univariate approaches such as beta calibration and isotonic regression. 
A simple solution here is to address the problem via calibrating the model empirically in a binary manner, which we discuss in the next section.

\section{The Empirical Approach: Adopting Logistic Calibration and Beta Calibration}
\label{sec:e_cal}

Our first proposed approach is to discretise the target variable and by this transform the regression task into a multi-class classification task. 
We can then apply two-class calibration methods in the one-vs-rest manner to obtain multi-class probability estimates, interpretable as a piecewise constant conditional density function for the original regression calibration task.

As in Equation~\ref{eq:reg_multi} and Definition~\ref{def:e_cal}, we first discretise the target variable by introducing $K$ segments defined by thresholds $-\infty=t_0<t_1<\dots<t_{K-1}<t_K=\infty$. 
Fitting of the calibration map for the regression model is performed as follows:
\begin{enumerate}
    \item For each class corresponding to one of the discretised segments $(t_{i-1},t_i]$ we build a training dataset for learning a one-vs-rest calibration model. Every instance $(\mathbf{x},y)$ in the calibration fold of the regression task is transformed into the estimated probability mass $G_{(\mathbf{x})}(t_i)-G_{(\mathbf{x})}(t_{i-1})$ and the binary ground truth label $\mathbb{I}(t_{i-1}<y\leq t_i)$.
    \item A binary calibration model $c_i:[0,1]\to[0,1]$ is trained separately on each class $i=1,\dots,k$ using the training data from step 1.
\end{enumerate}
The CDF $G_{(\mathbf{x})}$ output by the regression model on a test instance $\mathbf{x}$ is calibrated as follows:
\begin{enumerate}
    \item For each $i=1,\dots,k$ we calculate the predicted probability mass $p_i=G_{(\mathbf{x})}(t_i)-G_{(\mathbf{x})}(t_{i-1})$ that the regression model puts on segment $(t_{i-1},t_i]$
    \item We apply the one-vs-rest calibration maps $c_i$ on the respective predicted probabilities $p_i$ and renormalise the results to ensure they add up to one. The calibrated probability vector $(q_1,\dots,q_K)$ has thus probabilities $q_i=c_i(p_i)/\sum_{j=1}^K c_j(p_j)$.
\end{enumerate}

%


Within this method we can use any 2-class calibration methods. In the experiments we will use logistic calibration and beta calibration. Beta calibration is more appropriate here because the input to the calibration method is already in the probability range $[0,1]$, whereas the logistic calibration derived from Gaussian assumptions would be best on the full real-valued scale. However, for reference we have still decided to include logistic calibration into the experiments. We will refer to the corresponding regression calibration methods as \emph{e-logistic} and \emph{e-beta}, where \emph{e-} stands for \emph{empirical}.

%
%
%
An example with both predictions from e-logistic and e-beta is given in Figure~\ref{fig:result_0}.
Notice that the calibration map in the middle of the figure is drawn by putting the uncalibrated CDF as the horizontal coordinate and using the calibrated CDF as the vertical coordinate, hence we refer to it as marginal calibration map as it marginalises the effect of $t$.
As shown in the figure, the calibrated PDFs from e-logistic and e-beta are close to each other, and both show a bi-modal shape around the original estimated mean of the Gaussian.
In this particular case, the true value indeed falls into one of the modes.
The interpretation here is natural, while the predicted Gaussian distribution is optimised for least errors, its uni-modal assumption pushes it to lie around the mean of the training values.
Hence, by adopting calibration methods, we show that the estimated PDFs are capable of generating a non-parametric shape of the predictive distribution from the original Gaussian, which captures the distribution of under-estimated values and over-estimated values around the original Gaussian mean.

Here both empirical approaches can be seen as non-parametric as the number of parameters increases with the number of target values, but not with the size of the dataset.
Therefore both approaches take roughly linear time in the size of the dataset, and in the number of target values.

\section{GPC: Using Gaussian Processes for Calibration}
\label{sec:gp_cal}

While the empirical approaches are quick to apply, they can not provide a smooth estimation of the CDFs and PDFs on the target variable, hence giving limited information regarding the predicted distribution of the target.
As introduced previously, both Logistic calibration are Beta calibration are derived by assuming certain distributions on the predicted probabilities, which can then be optimised with a probabilistic objective function to approximate the calibrated probabilities.
Intuitively, it would be ideal if we can also make such distributional assumptions in the regression case.
Here we propose an approach based on the Gaussian Process Classifier (GPC) \cite{Williams1998,Rasmussen2006} to achieve a smooth calibration function, which can be seen as modelling a latent Gaussian Process over the CDFs. 

Following Definition~\ref{def:reg_cal}, to calibrate a regression model we need a calibration function in the following form, denoting $\mathsf{q}(t) = G_{(\mathbf{x})}(t)$:
\begin{align*}
    c(t, \mathsf{q}) = \hat{\mathsf{q}}(t) 
\end{align*}
%
As discussed above, in general we cannot design $c$ with a finite dimensional vector to represent $\mathsf{q}$, unless $\mathsf{q}$ follows certain parametric assumptions.
For instance, for the case of Gaussian, we can use the mean and standard deviation to represent the function.
However, as parametric assumptions can be a potential reason for yielding uncalibrated CDFs, here we strategically avoid such approaches.

Therefore, we consider a non-parametric approach which does not require an explicit representation of the whole function of $\mathsf{q}$, but only takes in a single value of $\mathsf{q}(t)$:
\begin{align*}
    c\big(t, \mathsf{q}(t)\big) = \hat{\mathsf{q}}(t) 
\end{align*}
We view this as a two-class probability estimation task with two features. The features are $t$ and $\mathsf{q}(t)$ and we want to predict the calibrated probability that the original regression target variable $Y$ is below the threshold $t$. To solve this task we use the Gaussian Process Classifier algorithm. 

First, we need to build the training set for GPC. For this we 
%
consider the set of $N$ predicted CDFs $(\mathsf{q}_1,...,\mathsf{q}_N)$ on the calibration fold instances, and a set of $K$ target values $(t_1,...,t_K)$.
The training instances are then $\mathbf{z}_{i,j} = (\mathsf{q}_i(t_j),t_j)$, representing a particular combination of the cumulative distribution $\mathsf{q}(t)$ and the corresponding value of $t$.
GPC models the probability estimator as a composition of two functions: a function $h$ which transforms the features into a hidden real-valued Gaussian-distributed variable encoding the confidence information, followed by a link function $\phi$ which transforms this confidence information into a probability.
That is, it models a function $h: [0, 1] \times \mathbb{R} \rightarrow \mathbb{R}$, assuming that the $N * K$ function values $h(\mathbf{z}_{1,1}), h(\mathbf{z}_{1,2}), \dots, h(\mathbf{z}_{N,K})$ are jointly Gaussian distributed, with a constant mean of $0$ and some $N*K$ by $N*K$ covariance matrix $\Sigma$.
Hence, instead fitting a distribution over the CDFs, we now have a distribution over the functions on a finite sample from the CDFs.

If we construct the covariance matrix $\Sigma$ via some covariance function $\mathbf{k}_{\theta}$ (a positive definite kernel with parameter $\theta$), so that $\Sigma_{m,n}^{(\theta)} = \mathbf{k}_{\theta}(\mathbf{z}, \hat{\mathbf{z}})$, the Gaussian distribution can be generalised to any infinite set of dimensions, which can later be used for making predictions.
The next step is to map the quantities of $h(\mathbf{z}_{i,j})$ into the interval of $[0,1]$, which can then be used to compute a objective function with the target variable $\mathbb{I}(Y \leq t)$.
The approach used in GPC is to adopt a link function $\phi: \mathbb{R} \times \{0, 1\} \rightarrow [0 ,1]$, which is commonly constructed using logistic function or probit function.
Training of GPC involves optimising the kernel parameter given $\mathbf{z}_{i,j} = (\mathsf{q}_i(t_i), t_j)$ and $b_{i,j} = \mathbb{I}(y_i \leq t_j)$, by marginalising out $h$:
\begin{align}
\label{eq:gpc_theta}
    \hat{\theta} = argmax_{\theta} \int_{h} \left(\prod_{i=1,j=1}^{N,K} \phi(h(\mathbf{z}_{i,j}), b_{i,j})\right) \cdot  Norm\Big(h(\mathbf{z}_{1,1}), \dots, h(\mathbf{z}_{N,K}) ; 0, \Sigma^{(\theta)}\Big) dh
\end{align}
Here $Norm()$ denotes the likelihood function of the multivariate Gaussian. 
Regarding how prediction works using the GPC model please refer to \cite{Rasmussen2006}.

As in common GPs, GPC is not sparse and hence has some computational difficulties.
The most widely adopted approximation is the Laplace approximation \cite{Williams1998,Rasmussen2006} and Expectation Propagation \cite{Minka2001}.
Both approaches are commonly seen in GP implementations as in scikit-learn \cite{scikit-learn}, GPy \cite{Gpy2014}, and Edward \cite{Tran2016}. 

\begin{figure}[t]
    \centering
    \includegraphics[width=0.45\textwidth]{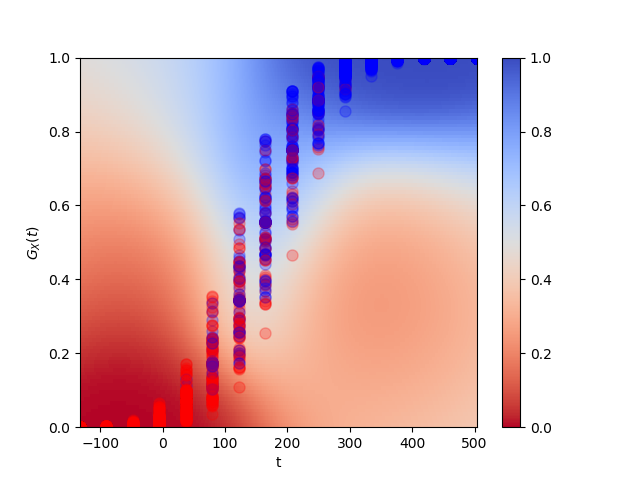}
    \includegraphics[width=0.45\textwidth]{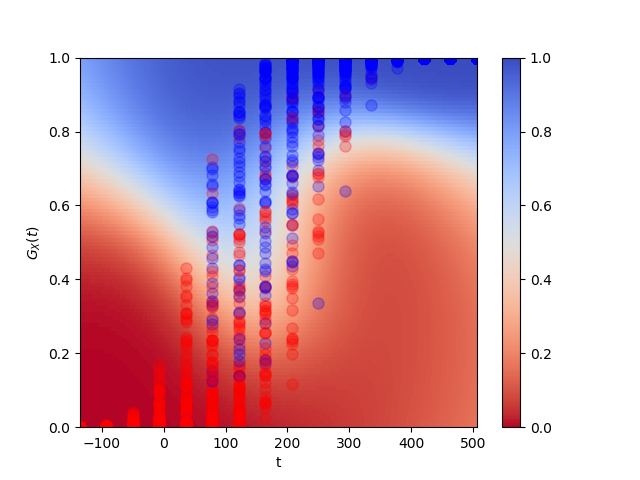}
    \caption{Two examples of the calibration map for the GPC approach using a RBF kernel, with base models outputting a Gaussian density. The blue and red points are corresponding to the training points of $\mathbb{I}(Y \leq t) = 1$ and $\mathbb{I}(Y \leq t) = 0$ respectively. 32 values of $t$ are selected uniformly.}
    \label{fig:gp_cal}
\end{figure}

To train a GPC calibration we first require a set of target values $t_1,...,t_K$, with which we can construct the input variable $\mathbf{z} = [\mathsf{q}(t), t]$, and the output variable $\mathbb{I}(Y \leq t)$ with the data points in the calibration set.
The next step is to train a GPC to predict $\mathbb{I}(Y \leq t)$ from $[t, \mathsf{q}(t)]$. 
While any positive kernel can be potentially applied, here we use the RBF kernel as a default option in many GP and SVM applications, given that our aim is to smoothly calibrate the CDFs with the provided training points.  
Two examples of the training points and estimated calibration map can be seen in Figure~\ref{fig:gp_cal}.

Since the GPC model is continuous, the thresholds $\hat{t}_1, ..., \hat{t}_{\hat{K}}$ do not need to be the same on training and test data. Therefore, on test data one can use many more thresholds than were used on the training data. 
While computationally we cannot select a infinity smooth set of $t$, this can be done empirically as a trade-off between precision and computational speed, as in general approximation approaches.
The following steps are again simple to perform. 
For a test feature $\mathbf{x}$ and the uncalibrated CDF $G_{(\mathbf{x})}(t)$, we again construct the input feature as $(t, G_{(\mathbf{x})}(t))$, and use the previous learned GPC to predict the estimated $\mathbb{\hat{P}}(Y \leq t \mid G_{(\mathbf{x})}(t))$. 
The estimated PDF can be then directly calculated as $\hat{g}_{(\mathbf{x})}(t_i) = \frac{\hat{G}_{(\mathbf{x})}(t_{i+1}) - \hat{G}_{(\mathbf{x})}(t_{i})}{t_{i+1}-t_{i}}$.

\begin{figure*}[t]
    \centering
    \includegraphics[width=0.95\textwidth]{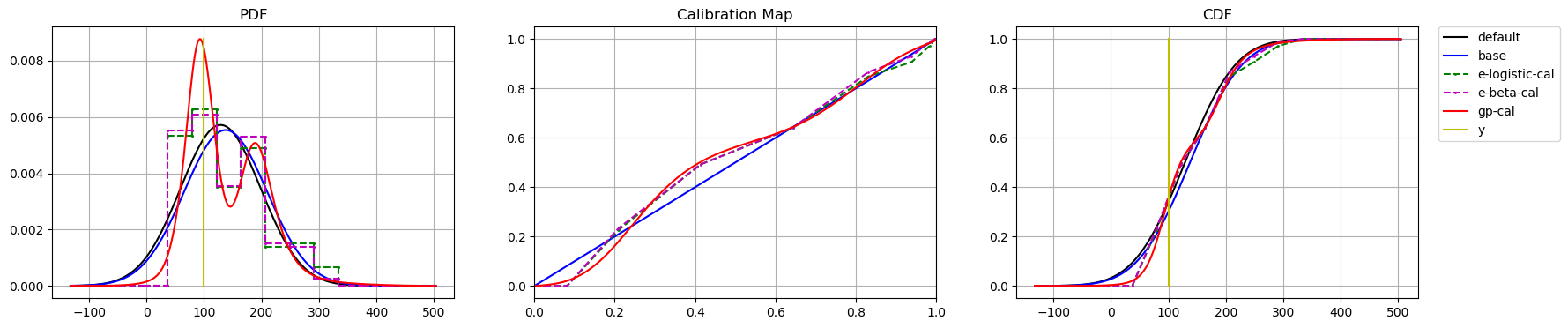}
    \caption{An example of the PDFs, marginal calibration maps, and CDFs estimated on a test instance using e-logistic, e-beta, and GPC. The base model is estimated with Gaussian Process Regression. Here default model refers to the model fitted with the whole training set, base models refers to the model fitted with $2/3$ of the training set, and the rest $1/3$ of the training set is used to learn the calibration (with $16$ linearly mapped target values). The PDFs are obtained by consecutively applying the base model and calibration maps on the test feature. The ground-truth is given as the yellow vertical line.}
    \label{fig:result_0}
\end{figure*}

A result of GPC calibration can be again seen in Figure~\ref{fig:result_0}.
As the figure indicates, GPC calibration captures a close bi-modal shape on the PDF as the ones of e-logistic and e-beta, but instead have a smooth estimation.
In this particular case, the smooth estimation provides a higher likelihood for the ground truth, and hence a lower log-loss, a major benefit of having calibrated outputs.

The major drawback of the GPC approach comes from its computational cost.
As in general GP approaches, the computation of a GPC require some numerical approximations involving the inverse of matrices.
This makes the speed of GPC relatively slow compared to the empirical approaches, and intractable for larger datasets (calibration sets), where further sparse approximations are required.

\section{Experimental Evaluation}
\label{sec:exp}

\begin{figure*}[t]
    \centering
    \begin{subfigure}{\textwidth}
    \centering
    \includegraphics[width=0.24\textwidth]{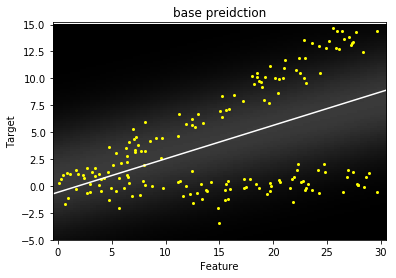}
    \includegraphics[width=0.24\textwidth]{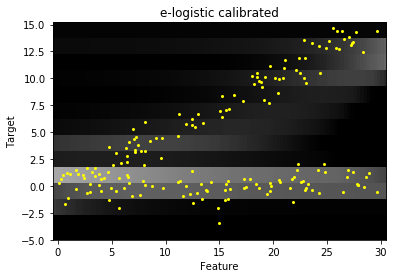}
    \includegraphics[width=0.24\textwidth]{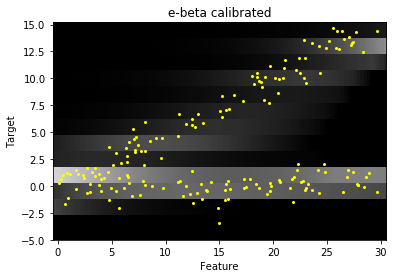}
    \includegraphics[width=0.24\textwidth]{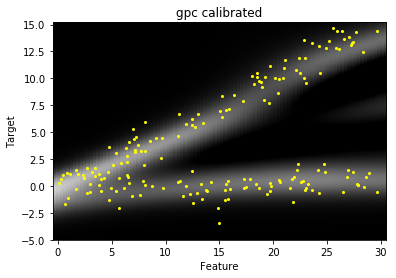}
    \caption{Bayesian Ridge Regression}
    \end{subfigure}
    \begin{subfigure}{\textwidth}
    \centering
    \includegraphics[width=0.24\textwidth]{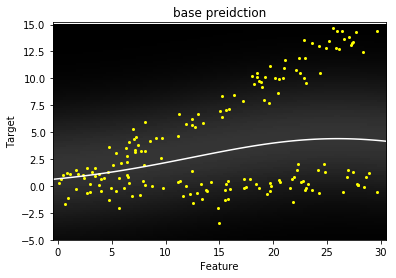}
    \includegraphics[width=0.24\textwidth]{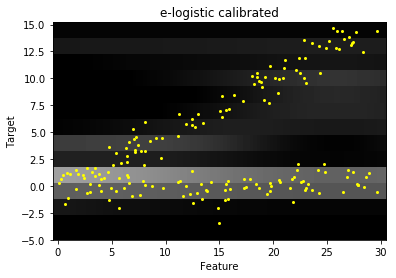}
    \includegraphics[width=0.24\textwidth]{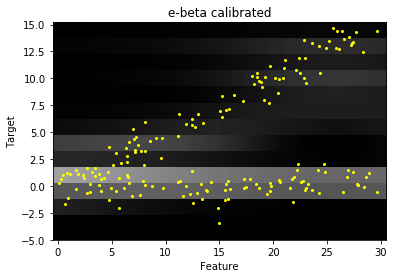}
    \includegraphics[width=0.24\textwidth]{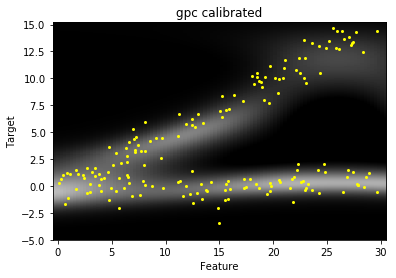}
    \caption{Gaussian Process Regression}
    \end{subfigure}
    \caption{Predicted densities on a toy dataset.The result on the left shows the densities predicted from training on $2/3$ of the dataset. The rest of the results are obtained by using the remaining $1/3$ of the data to train a calibration method, and then applying it upon the base model on the left. The white lines show the predicted mean from the corresponding regression models. For the calibration methods, 16 target thresholds with equal distance are applied on the y-axis, which hence provide 16 bins for the predictions from the empirical methods. For the GPC approach, while also training 16 target thresholds, at test time 256 target thresholds are further specified to generate a smooth output.}
    \label{fig:toy_results}
\end{figure*}

\begin{figure*}[t]
    \centering
    \begin{subfigure}{\textwidth}
    \centering
    \includegraphics[width=0.24\textwidth]{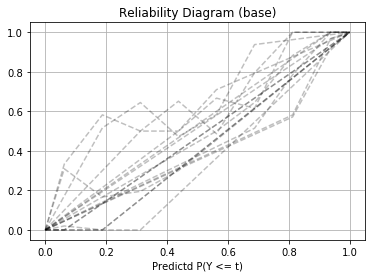}
    \includegraphics[width=0.24\textwidth]{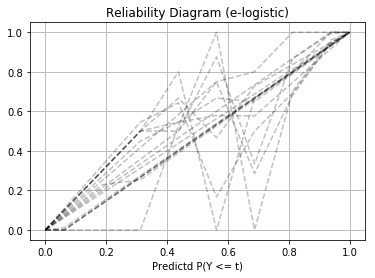}
    \includegraphics[width=0.24\textwidth]{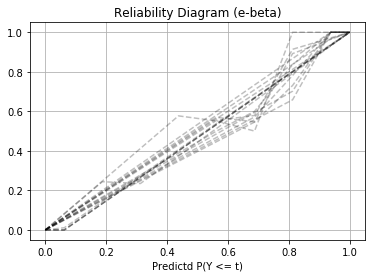}
    \includegraphics[width=0.24\textwidth]{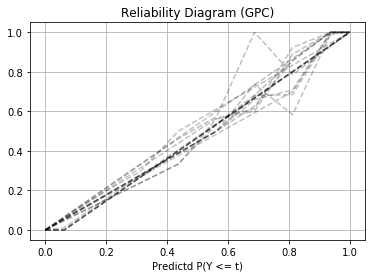}
    \caption{Bayesian Ridge Regression}
    \end{subfigure}
    \begin{subfigure}{\textwidth}
    \centering
    \includegraphics[width=0.24\textwidth]{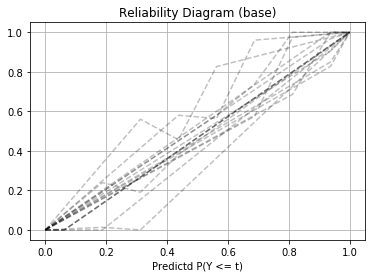}
    \includegraphics[width=0.24\textwidth]{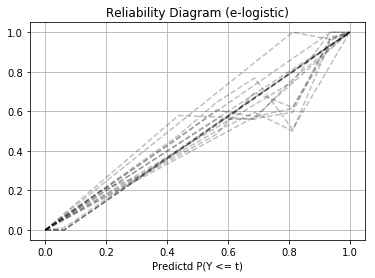}
    \includegraphics[width=0.24\textwidth]{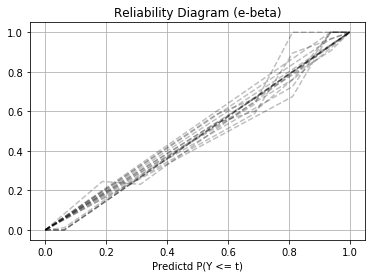}
    \includegraphics[width=0.24\textwidth]{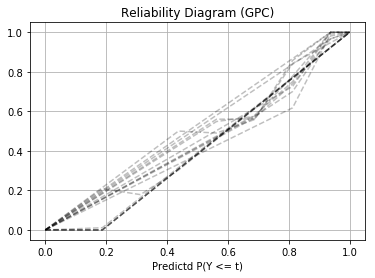}
    \caption{Gaussian Process Regression}
    \end{subfigure}
    \caption{Reliability diagrams on a toy dataset. Each dashed line is drawn by a particular target value $t$, with estimated probability for $Y \leq t$ on x-axis, and the actual relative frequency of $Y\leq t$ on the y-axis (with $8$ bins on the x-axis). }
    \label{fig:toy_rm}
\end{figure*}

In this section we experimentally examine the performance of our proposed methods, and compare them against different uncalibrated regression models.
We first revisit the toy dataset used at the beginning of the paper. 
We then use 5 UCI datasets to compare multiple regression models.

As base models we selected three methods with Gaussian outputs: Ordinary Least Squares regression (OLS), Bayesian Ridge Regression (BRR), and Gaussian Process Regression (GPR).
This choice is motivated by the following reasons.
(1) Gaussian-output models are the most common among probabilistic regression methods, and have been used as baseline approaches in most regression problems.
(2) These three models covers different aspects of a Gaussian-output method.
OLS is optimised by squared error, which is the equivalent of fitting a linear function to predict the mean of a Gaussian output with a shared standard deviation. 
While BRR is still a linear model, all its parameters are optimised through a posterior given certain priors (in this case uninformative priors are used, which acts as regularisers).
GPs can give non-linear predictions with certain kernels (in the following experiments RBF is used), and is optimised through a likelihood function.
However, as stated previously, our proposed approaches are not limited to Gaussian-output methods -- our main goal here is to compare performance among different model assumptions.     

In terms of implementation, for all experiments we apply the same experimental design as in \cite{Platt1999,Kull2017}, which runs 5-fold cross validation.
Given a base model class and a calibration method, a calibrated regression can be trained by separating the training set into a base set and a calibration set.
For each execution, the training set is divided into another 3 folds to iteratively train the base models and the calibration methods, which provides three calibrated models.
The base model is first fitted with the base set, and then used to provide predictions on the calibration set.
The calibration is then learnt on the calibration set with the these predictions from the base model.  
Finally, during testing, the predictions are obtained by applying the learnt base model and calibration consecutively, the final predictions are given as the averaged prediction among all three calibrated models.

\subsection{The Toy Dataset}

In Figure~\ref{fig:toy_example} we showed an initial example with OLS to demonstrate the motivation of applying calibration on regression tasks. 
Here we use the dataset again to compare our proposed methods and uncalibrated models.
The dataset is generated as a mixture of two lines with a given Gaussian noise, with uniformly generated features on the horizontal axis.
The 5-fold cross validation provides the following results.

Figure~\ref{fig:toy_results} visually shows the predicted densities from both BRR and GPR from a single training, with calibrated densities from them using e-logistic, e-beta, and GPC respectively.
We omit the results of OLS here as it is partly shown in Figure~\ref{fig:toy_example} and close to the results of BRR in this particular case.
In general, all three proposed approaches are able to capture the bi-modal shape of data distribution towards larger input values, and can correct the base output to be closer to the true distribution as given in Figure~\ref{fig:toy_example}.
Both e-logistic and e-beta clearly show horizontal density bands across the figures, which is expected given their empirical nature.
Notably, the calibrated results with GPR under-estimates the densities around the top right of the figure.
The explanation can be obtained by checking the original output of the GPR, which shows a non-linear estimation by virtue of the RBF kernel, and also under-estimates the densities at the same x-location in the top-right area.
As discussed previously, while calibration can help improve the probability estimates from a given model, it can not further correct the predictions that are already grouped together.
In this case, the non-linear GPR provides the same density estimation for the top-left area as many other low-density areas, meaning this area cannot be simply fixed by applying calibration.

Figure~\ref{fig:toy_rm} shows the reliability diagram obtained by evaluating the $16$ training target values for each experiment.
Reliability diagram is a widely adopted tool in binary classification for visualising whether a classifier gives calibrated probability estimates.
The idea is to apply a set of bins on the probability estimates.
Then within each bin, we calculate the averaged value of the estimates, as well as the relative frequency of the binary target.
Then if we draw the two values within a 2-D space, a calibrated classifier will stay close to the ascending diagonal. 
In probabilistic regression we can obtain a set of lines with each being drawn as a binary task with the binary indicator $\mathbb{I}(Y \leq t)$. 
Both base models can be seen to be uncalibrated for certain values of $t$, as there are multiple lines away from the ascending diagonal.
All the calibration methods illustrates improved performance with most lines close to the ascending diagonal.
The exception is the e-logistic approach with Bayesian ridge regression, where the approach created a few points further away from the diagonal.
This is explainable as by definition logistic calibration is not designed for calibrating probabilistic models, and can lead to uncalibrated estimates for certain datasets and models \cite{Kull2017,Kull2017b}.

\subsection{Experiments on UCI Data}


\begin{figure*}[!ht]
    \centering
    \includegraphics[width=0.3\textwidth]{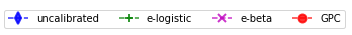}
    \begin{subfigure}{\textwidth}
    \centering
    \includegraphics[width=0.18\textwidth]{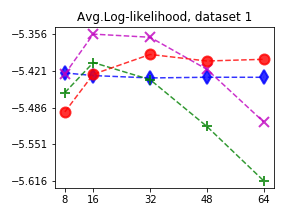}
    \includegraphics[width=0.18\textwidth]{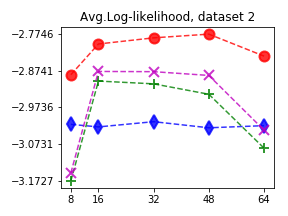}
    \includegraphics[width=0.18\textwidth]{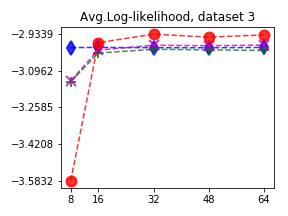}
    \includegraphics[width=0.18\textwidth]{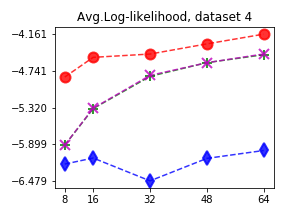}
    \includegraphics[width=0.18\textwidth]{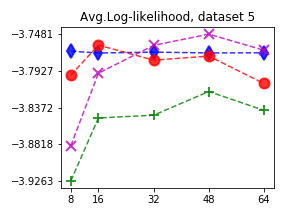}
    \caption{Ordinary Least Squares regression}
    \end{subfigure}
    \begin{subfigure}{\textwidth}
    \centering
    \includegraphics[width=0.18\textwidth]{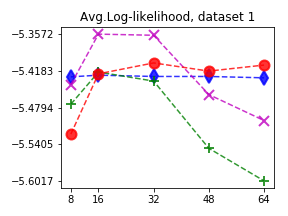}
    \includegraphics[width=0.18\textwidth]{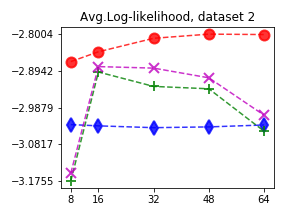}
    \includegraphics[width=0.18\textwidth]{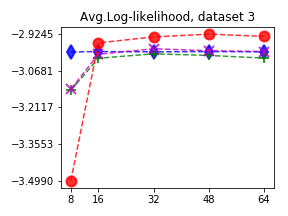}
    \includegraphics[width=0.18\textwidth]{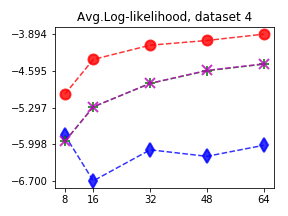}
    \includegraphics[width=0.18\textwidth]{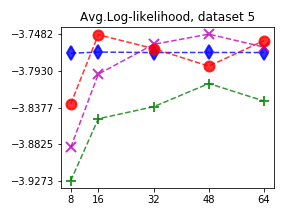}
    \caption{Bayesian Ridge Regression}
    \end{subfigure}
    \begin{subfigure}{\textwidth}
    \centering
    \includegraphics[width=0.18\textwidth]{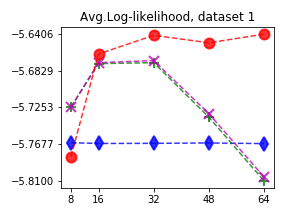}
    \includegraphics[width=0.18\textwidth]{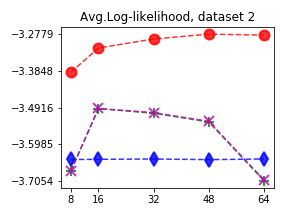}
    \includegraphics[width=0.18\textwidth]{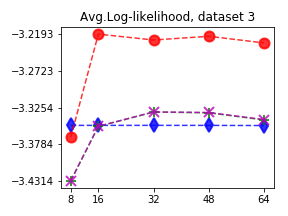}
    \includegraphics[width=0.18\textwidth]{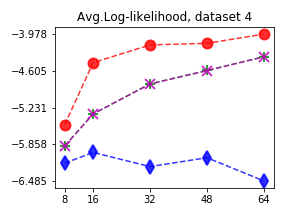}
    \includegraphics[width=0.18\textwidth]{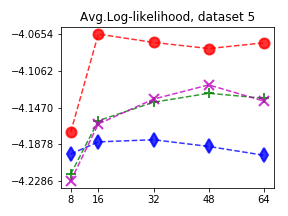}
    \caption{Gaussian Process Regression}
    \end{subfigure}
    \caption{Experiments with 5 UCI datasets. The x-axis indicates the number of target values used for training of calibration method. The y-axis shows the log-likelihood for the final estimate $\hat{p}(Y=y\mid \mathbf{X} = \mathbf{x})$, higher value indicate better results. Each column of figures is corresponding to one of the five UCI datasets. }
    \label{fig:uci_results}
\end{figure*}

While in the previous experiment we used artificial data to demonstrate a case where the true distribution is known, this experiment aims to investigate the performance of our methods with real datasets.
We use the log-likelihood as the evaluation measure for our experiments as is common for predictive probabilistic approaches.

We selected five datasets from the UCI repository \cite{Lichman2013}: (1) Diabetes, (2) Boston, (3) Airfoil, (4) Forest Fire, (5) Compressive Strength.
These five datasets are selected according to their size and formats.
We selected the size to be no more than $2\,000$ considering the speed of the GPC approach. 
Also, as later shown, we perform experiments to examine different numbers of target values, which is also time-consuming even on a single dataset.
Regarding the formats, we selected datasets that have a single tabular file and contain ready-to-use feature and target instances, which makes the experiments simple to reproduce.
The only pre-processing applied is to remove instances with missing feature values.

The experiments are organised as follows.
At the top level, as in \cite{Platt1999,Kull2017}, we run 10-times 5-fold cross-validation to provide the averaged results.
For the experiments with GPR as the base model we only use a single feature with the largest variance to ensure the convergence of the optimised kernel parameters.
At a detailed level, for all the calibration approaches, we select different numbers ($8, 16, 32, 48$ and $64$) of target values with equal distances, which then aims to test the effect the number of target values.
The prediction of GPC is set to have $1\,024$ target values, again with equal distance among neighbouring values.
The range of the target values is selected as $[y_{min} - 0.5 \cdot y_{range}, y_{max} + 0.5 \cdot y_{range}]$, where $y_{min}$ and $y_{max}$ are the minimum value and maximum value of the target variable in the training set, and $y_{range} = y_{max} - y_{min}$.
This setting ensures the estimated PDFs can approximately cover most of the probability mass (hence the CDFs can be approximately seen as in $[0,1]$).
To maintain the speed of the GPC approach, we use up to $5\,000$ CDF values from the base model, which are uniformly selected from all the outputs within the calibration set.

The results are depicted in Figure~\ref{fig:uci_results}.
Although the performance of our proposed methods can vary in different settings, it can be seen that there is always a calibrated method giving better estimation than the uncalibrated models.
The exception is the setting with the smallest number of target values ($8$, on the left), where the calibration methods mostly perform poorer than the uncalibrated ones.
This is reasonable as we only provide limited information from the CDFs to the calibration methods in this case.

With the empirical approaches, both e-logistic and e-beta outperforms the uncalibrated models while the number of the target values is around $16$ and $32$, and the performance tends to drop as the number becomes larger.
This drop can be explained by their empirical nature, where more empirical measurement can increase the variation of the output, hence increasing the potential for over-fitting. 
Furthermore, e-beta shows a better result than e-logistic for most cases.
This is expected as e-beta is able to give estimates beyond the Sigmoid function, which is shown to be more suitable for probabilistic calibration, as shown in \cite{Kull2017b}. 

For most datasets and settings the GPC approach achieves top performance, mostly benefitting from a larger number of target values.
However, several drops in performance can still be seen while $64$ target values are used.
This can be considered as a consequence of setting the $5\,000$ CDF values during the training process, which is equivalent to applying a naive sparse GPC, ending up with faster training but worse performance.

\section{Conclusion and Future Work}
\label{sec:con}

We investigated the problem of calibrating a probabilistic regression model to provide better probability estimates.
Compared to switching or improving the regression model itself, calibration provides an alternative approach to improve the original model directly. 
While we first define the concept of calibration in regression, we further illustrate that calibrated cumulative distribution predictions can lead to calibrated density predictions. 
One benefit of calibrating a model with CDFs is that we no longer require a parametric assumption on the density functions, which is useful if the distribution of the target is unknown.
Two empirical approaches are proposed based on Logistic calibration and Beta calibration. 
These approaches are useful if one wants to quickly calibrate the shape the predicted densities, without caring about a particular density value, or cumulative density value. 
We further propose an approach based on the Gaussian process classifier, which can learn a smooth calibration function on the predicted cumulative densities.
While the non-sparse property makes the approach relatively slow to train and not scale with larger datasets, it is useful for the scenarios where calibrated cumulative densities are required for decision making, such as forecasting tasks in areas like medicine. 

While we mainly investigate non-parametric methods given their versatility in the regression setting, parametric methods are still an alternative direction which is useful when the distribution of the target is indeed known, or can be approximated with reasonable uncertainty.
Among our proposed approaches, the empirical approaches are currently implemented via one-vs-rest, where further strategies can be investigated to provide improved estimations, such as the Least Square Error-Correcting Output Codes (LS-ECOC) approach proposed in \cite{kong1997probability}.
GPC can be developed further to incorporate large datasets, which can be linked to recent progress in the area of sparse Gaussian processes.


\section{Acknowledgements}

This work was supported by the SPHERE Interdisciplinary Research Collaboration, funded by the UK Engineering and Physical Sciences Research Council under grant EP/K031910/1.
MK was supported by the Estonian Research Council under grant PUT1458.

\bibliographystyle{splncs04}
\bibliography{main} 

\end{document}